\documentclass{bvm}

\makeatletter
\let\blx@rerun@biber\relax
\makeatother

\begin{document}
\selectlanguage{english}

\title{Internal Organ Localization Using Depth Images}
\subtitle{A Framework for Automated MRI Patient Positioning}

\author{
	Eytan \lname{Kats} \inst{1}, 
	Kai \lname{Geißler} \inst{2}, 
	Jochen~G. \lname{Hirsch} \inst{2}, 
	  Stefan \lname{Heldman} \inst{2}, 
	Mattias~P. \lname{Heinrich} \inst{1}
}
\authorrunning{Kats et al.}
\institute{
\inst{1} Institute of Medical Informatics, University of Lubeck\\
\inst{2} Fraunhofer Institute for Digital Medicine MEVIS, Bremen\\
}
\email{eytan.kats@uni-luebeck.de}

\maketitle

\begin{abstract}
Automated patient positioning is a crucial step in streamlining MRI workflows and enhancing patient throughput. RGB-D camera-based systems offer a promising approach to automate this process by leveraging depth information to estimate internal organ positions. This paper investigates the feasibility of a learning-based framework to infer approximate internal organ positions from the body surface. Our approach utilizes a large-scale dataset of MRI scans to train a deep learning model capable of accurately predicting organ positions and shapes from depth images alone. We demonstrate the effectiveness of our method in localization of multiple internal organs, including bones and soft tissues. Our findings suggest that RGB-D camera-based systems integrated into MRI workflows have the potential to streamline scanning procedures and improve patient experience by enabling accurate and automated patient positioning. 
\end{abstract}

\section{Introduction}
\label{3660-introduction}

The growing demand for Magnetic Resonance Imaging (MRI) examinations underscores the need to optimize scanning procedures to reduce patient wait times and improve operational efficiency. A critical, yet time-intensive, aspect of this process is the patient preparation and planning phase, which is essential for optimizing image quality and diagnostic accuracy \cite{van2020cinderellas}. This phase typically involves manual positioning of the patient followed by scout imaging to facilitate detailed geometric planning for subsequent diagnostic scans \cite{koken2009towards}. Automated patient positioning systems, capable of accurately identifying and localizing the anatomical region of interest, offer the potential to streamline this process by automatically adjusting the patient table to the correct position. Such systems could substantially reduce the time spent on manual adjustments, minimize positioning errors, and enhance workflow efficiency within radiology departments.

RGB-D camera-based systems have emerged as innovative tools in radiology workflows, particularly in enhancing patient setup and positioning processes. These systems provide precise mapping of the patient's external body surface. Studies by Incetan et al. \cite{incetan2020rgb} and Karanam et al. \cite{karanam2020towards} demonstrate the effectiveness of these systems in automated patient positioning by localizing key anatomical landmarks on the body surface, such as the head, thorax, and knee. Researchers have shown that predicting internal anatomical structures from the outer surface is also possible. Wu et al. \cite{wu2018towards} proposes a learning-based framework to generate a volumetric phantom, including the skeleton and lungs, from a detailed mesh representation of the patient's body surface. Recent line of works \cite{guo2022smpl, keller2022osso, shetty2023boss} integrate shape representation of internal organs into the popular Skinned Multi-Person Linear (SMPL) model \cite{loper2015smpl} that provides a flexible statistical framework to capture human pose and shape deformations. The SMPL-A method \cite{guo2022smpl} estimates the deformation of organs as a patient moves, but the organs’ initial shapes are obtained by a scan of the patient. OSSO \cite{keller2022osso} and BOSS \cite{shetty2023boss} demonstrate the ability to infer skeletal and organ shape from the body surface.  Although these methods show promising results, they rely on initial body surface mesh generation, which introduces additional processing time. Moreover, SMPL-based approaches require fitting the acquired surface mesh to obtain accurate model parameters, adding a further step to the inference process. In contrast, Teixeira et al. \cite{teixeira2023automated} propose a method that leverages depth camera images to directly estimate key patient scanning parameters, including lung position, isocenter, and water-equivalent diameter.

We investigate the feasibility of a learning-based framework to estimate approximate internal organ positions using simulated depth images derived from MRI scans. By leveraging depth information, the trained model infers organ locations, enabling the dynamic adjustment of patient table positioning and scan parameters, offering a potential solution to streamline radiological workflows. To ensure broad generalizability, we train our model on a large-scale dataset of approximately 10,000 full-body patient scans from the National German Cohort (NAKO) dataset \cite{bamberg2022whole}. Our experiments demonstrate the efficacy of the proposed solution in estimating the positions of 11 internal anatomical structures, including bones and soft tissue organs.

Our main contributions are as follows: (1) We demonstrate the efficacy of a learning-based framework in estimating internal organ locations solely from depth images. (2) We showcase the model’s generalizability across diverse body types and anatomical variations through training on a large-scale dataset.

\section{Materials and methods}
\label{3660-materials_and_methods}

In our study, we utilize approximately 10,000 full-body MRI scans from the NAKO dataset \cite{bamberg2022whole}. This extensive collection of images enables us to model anatomical information across a diverse range of anatomical variations within the population.

To simulate depth images, we normalize the MRI images to a range of 0 to 1 and apply a threshold of 0.02 to obtain a binary mask. Subsequently, we employ binary morphological opening to remove noise. Next, we extract a 2D orthographic depth image in the coronal plane by isolating the skin surface. After normalizing the depth image to a range of 0 to 1, we set to 0 all values smaller than 0.3 to eliminate table artifacts that are located far from the simulated camera position. Finally, we further denoise the resulting grayscale image using grayscale morphological opening. The resulting images are visualized in left column of Fig.\ref{3660-segmentation}.

To ensure a representative evaluation dataset, we selected 30 cases based on criteria such as study center, patient sex, age, weight, and height. Three radiologic technologists created segmentation masks for multiple structures within these cases. To obtain the segmentation masks for training, we employed the TotalSegmentator version for MRI images \cite{d2024totalsegmentator} to process the remaining scans. We then generated binary 2D masks by projecting the volumetric segmentations of each organ onto the coronal plane. Each organ was projected separately to account for potential overlap in the 2D plane, resulting in distinct binary masks for each structure. These masks encompass 11 internal organs, including three bone structures (hips, femurs, and a vertebra), two thoracic organs (heart and lungs), and six abdominal organs (kidneys, liver, pancreas, spleen, stomach, and urinary bladder).

\begin{figure}[b]
\centering
\includegraphics[width=\textwidth]{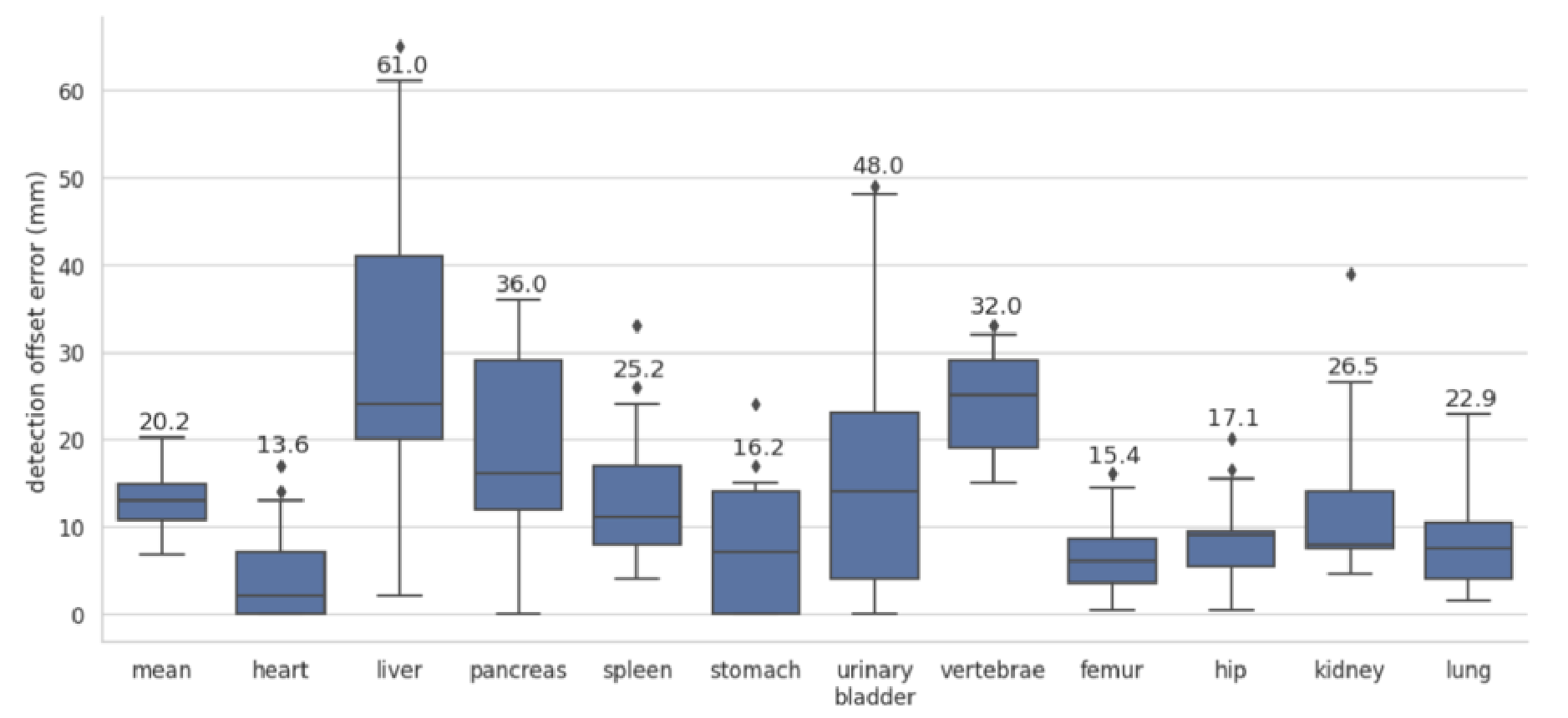} 
\caption{Detection Offset Error (DOE) for anatomical structures in 30 manually segmented MRI scans. The value above each box corresponds to 95 percentile of DOE.}
\label{3660-boxplot}
\end{figure}

For training, we employ a U-Net-based architecture that takes generated depth images of the body surface as input and outputs segmentation masks of internal organs. Ground truth segmentation masks are generated using TotalSegmentator. We train our model in a multi-label setting using a sigmoid activation function. We utilize a weighted combination of Dice loss and binary cross-entropy loss, with equal weighting of 0.5. During training, we employ a batch size of 16 and the Adam optimizer with an initial learning rate of 0.002, which is decayed using a cosine annealing schedule.

\section{Results}
\label{3660-results}

We evaluate the performance of the trained models on two distinct sets of ground truth segmentation masks. The first set comprises 30 images manually labeled by expert radiologic technologists, providing high-quality annotations. The second set is a hold-out collection of 696 masks generated by TotalSegmentator, similar to those used for the model training. Evaluation on manually labeled data assesses the model's ability to align with the 'gold standard' ground truth. Testing on additional automatically generated masks reveals its consistency with the training data and helps identify any biases introduced by the automated annotation process. The dual evaluation helps to assess the model's generalization capabilities and identify potential domain shifts between manually labeled and automatically generated data.

 To evaluate the model's ability to accurately predict organ shape, we compute the Average Symmetric Surface Distance (ASSD) between predicted and actual organ boundaries. We use the Dice coefficient to assess segmentation performance. Additionally, we derive bounding boxes from the segmentation masks and calculate the Detection Offset Error (DOE), defined as the distance between the corresponding sides of the ground truth and predicted bounding boxes. For each detected organ, we use the maximum value among the top, bottom, left, and right offsets in our calculations. This metric is particularly relevant for scanning procedures, as it quantifies the margin needed for precise patient table positioning to ensure complete coverage of the target organ during scanning. Chosen metrics provide a comprehensive evaluation of both localization accuracy and shape fidelity, enabling an assessment of the model's clinical potential.

We investigate the correlation between external body surface and internal organ location and shape. To this end, we train our model on incrementally larger datasets, tracking performance improvements for 11 anatomical structures (Tab.\ref{3660-results_comparison}). As expected, model performance improves with increasing training data, with each incremental improvement confirmed as statistically significant by a Wilcoxon signed-rank test when $p < 0.0001$. However, surprisingly, even with only 100 training examples, the model can capture variations in organ location and shape across the population.

\begin{table}[t]
\caption{The model's performance for internal organ segmentation from body surface depth images is evaluated using the mean value of the following metrics across 11 anatomical structures: Dice coefficient, Average Symmetric Surface Distance (ASSD), and percentile 95 of Detection Offset Error (DOE). Results are reported for a manually labeled data and an automatically generated masks using TotalSegmentator (TS).}
\label{3660-results_comparison}

\resizebox{\textwidth}{!}{
\begin{tabular*}{\textwidth}{c@{\extracolsep\fill}cccccc}
\hline

\# of training & \multicolumn{2}{c}{Dice} & \multicolumn{2}{c}{ASSD (mm)} & \multicolumn{2}{c}{DOE (mm)} \\
examples & manual & TS & manual & TS & manual & TS \\

\hline

100 & 72.44 $\pm$ 7.01 & 75.66 $\pm$ 6.4 & 10.11 $\pm$ 3.33 & 9.66 $\pm$ 3.32 & 48.95 & 40.0 \\
1000 & 76.01 $\pm$ 4.42 & 78.82 $\pm$ 5.03 & 8.93 $\pm$ 1.74 & 8.87 $\pm$ 3.34 & 42.33 & 31.0 \\
10000 & 79.44 $\pm$ 3.55 & 81.46 $\pm$ 4.52 & 7.94 $\pm$ 1.49 & 8.5 $\pm$ 3.43 & 36.0 & 29.0 \\

\hline
\end{tabular*}}
	
\end{table}

\begin{figure}[b]
\centering
\includegraphics[width=\textwidth]{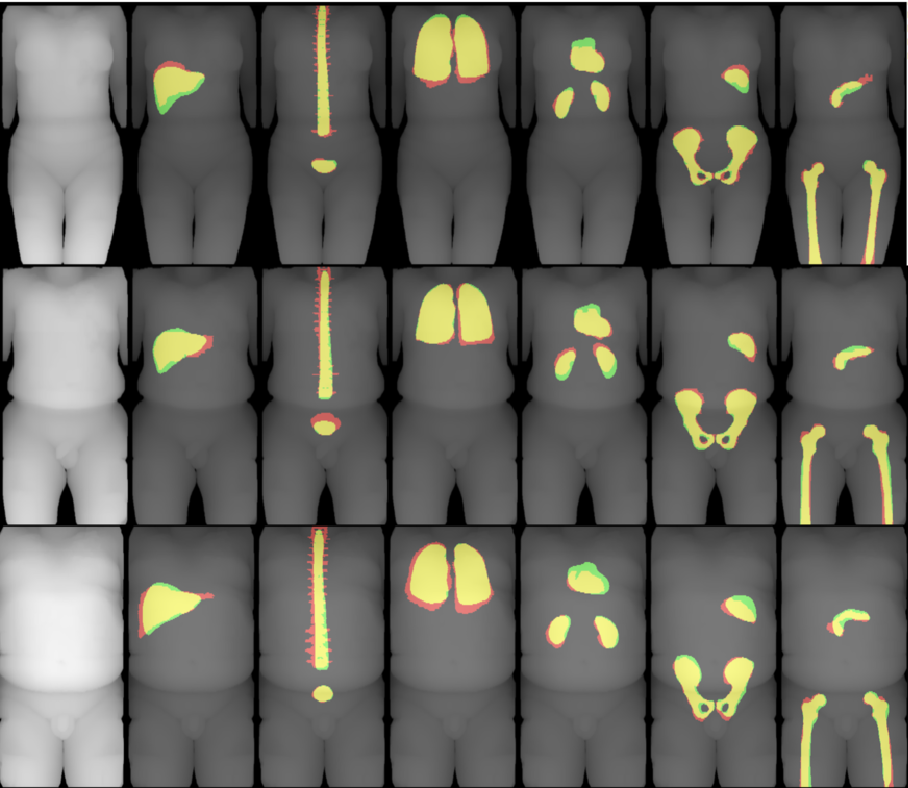} 
\caption{Comparison of predicted and ground truth segmentations for internal organs. To enhance visualization clarity and address potential overlaps of 2D projections, organs are presented in a separate images. From left to right: depth image derived from MRI scan, liver, vertebrae and urinary bladder, lungs, heart and kidneys, spleen and hips, femurs and pancreas. Red - ground truth mask, green - model prediction, yellow - overlap between ground truth and predicted mask.}
\label{3660-segmentation}
\end{figure}

We observed a performance gap between the model's performance on manually annotated data and automatically generated data from TotalSegmentator (Tab. \ref{3660-results_comparison}). However, this gap is relatively small, suggesting that automatically generated masks can be effectively utilized to train robust models for clinical applications. 

Quantitative analysis based on the DOE demonstrates the model's ability to accurately predict internal organ location (Fig.\ref{3660-boxplot}). For most structures, the 95th percentile of the DOE is below 30mm. However, precise localization of the urinary bladder (Fig.\ref{3660-segmentation} middle row) and left liver lobe tip (Fig.\ref{3660-segmentation} bottom row) remains challenging due to variability across population.

Qualitative analysis reveals the model's ability to infer organ shapes, with particularly intriguing results for soft-tissue organs (Fig.\ref{3660-segmentation}). Notably, the model captures natural variations in liver shape and accurately localizes the kidneys, despite their relative position variations across individuals.

These findings suggest that body surface depth information can provide valuable insights into internal organ anatomy, demonstrating the potential of integrating camera-based organ localization into radiological workflows.

\section{Discussion}
\label{3660-discussion}

In this study, we present an approach to automate patient positioning using depth images. Our results demonstrate the feasibility of using depth images to estimate organ positions and shapes. The model's ability to accurately predict organ locations suggests its potential to streamline the patient setup process and reduce the time required for manual adjustments, thereby improving the overall efficiency of MRI examinations.

However, it's important to note that the depth images used in this study were derived from MRI scans, which may not perfectly represent real-world depth camera images. Real-world images can exhibit variations in lighting conditions, noise, patient movement and patient clothing, potentially impacting the model's performance. Addressing this domain gap between simulated and real-world data is a crucial area for future research.

Furthermore, exploring the potential of 3D organ shape prediction and localization can further enhance the accuracy and clinical utility of automated patient positioning systems. By predicting the 3D shape and position of organs, we can gain a more comprehensive understanding of their spatial relationships and improve the precision of patient positioning.

\begin{acknowledgement}
We gratefully acknowledge the financial support by German Research Foundation: DFG, HE 7364/10-1, project number 500498869.
\end{acknowledgement}

\printbibliography

\end{document}